\documentclass{article} 
\usepackage[final]{colm2026_conference}

\usepackage{microtype}
\usepackage{hyperref}
\usepackage{url}
\usepackage{booktabs}
\usepackage{graphicx}
\usepackage{multirow}
\usepackage{amsmath}
\usepackage{xcolor}
\usepackage{url}
\usepackage{soul}

\usepackage{lineno}

\definecolor{darkblue}{rgb}{0, 0, 0.5}
\hypersetup{colorlinks=true, citecolor=darkblue, linkcolor=darkblue, urlcolor=darkblue}

\author{
  Jane Adkins$^{1}$, Abigail Walsh$^{1}$, Brian Davis$^{1}$, Elaine Uí Dhonnchadha$^{2}$ \\
  $^{1}$ADAPT Centre, Dublin City University \\
  $^{2}$Trinity College Dublin \\
  \texttt{$^{1}$firstname.lastname@adaptcentre.ie}
}

\begin{document}

\ifcolmsubmission
\linenumbers
\fi
\title{MoirfEolas and CríochScore: Developing Resources for and the Evaluation of Tokenization Alignment with Irish Morphology}
\maketitle

\begin{abstract}
This paper presents new tokenization resources for Irish and evaluation measures of alignment with the morphological boundaries of the language. We present \texttt{MoirfEolas}, a dataset of over 35,000 Irish words mapped to their respective eclipses, prefixes and suffixes as well as an evaluation metric \texttt{CríochScore}, that evaluates the alignment of tokenizations with the morphological boundaries present in \texttt{MoirfEolas}. We evaluate common tokenization algorithms using \texttt{CríochScore} as well as intrinsic metrics present in the tokenization literature. We find that the Unigram Language Model aligns with Irish morphology more often than the other algorithms evaluated. We also find trade-offs between morphological-alignment of tokenization with both compression as well as vocabulary efficiency, providing practical insights for Irish natural language processing development. This dataset contributes towards combating the Irish language's low-resource status; moreover, the construction process reported in this paper can be emulated by other languages to create specialised morphological resources.
\end{abstract}

\section{Introduction}
Tokenization is the process of breaking data down into discrete units referred to as tokens and forms a crucial preprocessing step in preparing data for input into large language models. Despite the lack of attention shown to this fundamental step, choice of tokenization algorithm has been shown to affect downstream performance of models {\citep{ali-etal-2024-tokenizer}}. Current subword-based approaches to tokenization have become the de-facto approach; these algorithms break text down into characters, subwords, or whole words, enabling the modeling of rare and out-of-vocabulary words.

Morphologically complex languages exhibit high degrees of lexical variation which may not be adequately captured by tokenizer vocabularies. Research into tokenization for such languages often explores the alignment of tokens with morphological boundaries \citep{arnett-bergen-2025-language, arnett2025evaluatingmorphologicalalignmenttokenizers} with conflicting findings as to whether alignment affects language modeling performance \citep{arnett-bergen-2025-language}; while morphological pre-tokenization has shown to improve performance downstream \citep[inter alia]{osvth-etal-2026-impact, hou-etal-2023-effects, erkaya-comprehensive-analysis-2022}, results reported by \citet{vemula2025rethinkingtokenizationrichmorphology} show that tokenizer choice has a greater effect on downstream performance than morphological alignment of tokens. 

Tokenization also has an effect for low-resource languages where the limited data available creates a '\textit{vocabulary bottleneck}' in multilingual language models \citep{chai-etal-2024-tokenization}. This bottleneck creates tokenization length disparities where underrepresented languages are encoded at greater lengths, with more and smaller tokens being used to represent them than for higher-resourced languages. This disparity snowballs into issues surrounding the context window, cost, processing times and storage requirements in language modeling \citep{petrov-unfairness-2023, arnett-etal-2024-bit}. 

We create resources for the evaluation of the morphological alignment of tokenization for Irish, a low-resource and morphologically complex language. Evaluation of commonly-used subword tokenization algorithms is conducted using these resources as well as intrinsic metrics proposed in the tokenization literature. Irish is the first official and national language of the Republic of Ireland. It displays high amounts of inflection--- most commonly through suffixation \citep{ui-dhonnchadha-2002-two}. Initial mutations are a feature of Irish through \textbf{eclipsis}, where a consonant is added to the beginning of a word (see Appendix \ref{app:eclipses} for details), and \textbf{lenition}, where a `h' is inserted after the initial character of a word\footnote{Lenition is not investigated as this process mutates the morpheme within a word}. We create the \textbf{MoirfEolas} dataset, mapping Irish words to their respective eclipses, prefixes and suffixes. As well as this, we present \textbf{CríochScore}, an intrinsic tokenization evaluation metric that ascertains the morphological alignment of tokenization segmentations with Irish morphology. Both the dataset creation and evaluation methodology are designed specifically for the Irish language, but can be easily adapted to other languages\footnote{All code for replicating dataset creation and morphological evaluation is available here: \url{https://github.com/jndkns01/tknzn4Irish}}. This work contributes to tackling the low-resource status of the Irish language, ultimately supporting the Digital Plan for Irish \citep{digitalplan}, a government strategy for improving language technologies for the nation's official language. Moreover, this work addresses a gap in Irish natural language processing research by evaluating the fundamental step of tokenization and how this step handles the unique features of the Irish language and its morphology.

\section{Background}
\subsection{Tokenization Algorithms}
Subword tokenization algorithms have largely replaced word-based approaches to tokenization. We outline the following prevalent tokenization algorithms, along with models that implement them.

\textbf{BPE} \citep{sennrich-etal-2016-neural} is based on a compression algorithm and utilises a merge-based vocabulary construction process. The algorithm begins with a base vocabulary of each of the distinct characters in the training corpus and merges commonly co-occuring characters until the pre-defined vocabulary size is reached. It is used in GPT\footnotemark{} and Galactica \citep{taylor2022galacticalargelanguagemodel}. \textbf{Byte-level BPE} \citep{Radford2019LanguageMA} is an augmented BPE process that operates on raw bytes instead of characters, used in GPT-2, 3, 4, 4o, code-davinci-001/002, text-davinci-003, gpt-3.5-turbo\footnotemark[\value{footnote}], RoBERTa \citep{liu2019robertarobustlyoptimizedbert} and BART \citep{lewis2019bartdenoisingsequencetosequencepretraining}. \textbf{WordPiece} \citep{schusternakijama2012wordpiece} is also a merge-based algorithm that selects the co-occuring characters that maximise the likelihood of the training data, i.e. it merges characters with maximum mutual information value, using an n-gram language model to score the likelihood of possible pairs. It is used by BERT \citep{devlin-etal-2019-bert} and ELECTRA \citep{clark2020electrapretrainingtextencoders}.
\footnotetext{\url{https://openai.com/}}

The \textbf{Unigram Language Model} (LM) is a pruning method that is initialised with a large base vocabulary from a training corpus. This is iteratively pruned by calculating the likelihood increase after removal of a subword unit, using the expectation maximisation algorithm to compute the probability of each unit to optimise the vocabulary. This pruning process ceases once the pre-defined vocabulary size is reached \citep{kudo-2018-subword}. 

\textbf{SentencePiece} is a library for tokenization. Unlike other tokenization algorithms, SentencePiece treats the input text as a raw sequence of Unicode characters with many default pre-tokenization steps such as splitting on whitespace, splitting by number, splitting by Unicode script which can be switched off in application. This design enables the construction of an end-to-end tokenization system that is language-independent and does not rely on any language-specific preprocessing. SentencePiece implements both BPE and Unigram LM algorithms \citep{kudo-richardson-2018-sentencepiece}. The BPE-implementation is used by LaMDA \citep{thoppilan2022lamdalanguagemodelsdialog}, Gemma, Gemma 2 \citep{gemmateam2024gemmaopenmodelsbased}, Llama and Llama 2 \citep{touvron2023llamaopenefficientfoundation}. The Unigram LM-implementation is used by XLNet \citep{yang2020xlnetgeneralizedautoregressivepretraining}, ALBERT \citep{lan2020albertlitebertselfsupervised} and T5 \citep{raffel2023exploringlimitstransferlearning}.

\subsection{Tokenization of Irish}
Research into tokenization for Irish has remained word- and rule-based rather than subword-based. Finite-state tokenization and morphological analysis tools were designed with special handling of multi-word expressions and contractions, to effectively preserve their semantic and syntactic coherence \citep{ui-dhonnchadha-2002-two, Kilgarriff2006, Cassidy2023} but remain word-based. Work on Old Irish has concentrated on harmonising annotation schemes and resolving tokenization inconsistencies across corpora through rule-based and character-level approaches \citep{doyle-etal-2019-character, doyle-mccrae-2025-assessment}. 

Subword tokenization has been discussed in the development of a transformer model for the Irish language, with algorithm choice being described as `\textit{the biggest driver of translation performance}' \citep{lankford-etal-2021-transformers}. As well as this, in the development of a BERT-based model for the Irish language---gaBERT---experimentation was done with SentencePiece\footnote{Unspecified whether the BPE or Unigram LM SentencePiece implementation} and WordPiece vocabularies, with both Irish and English data to account for the use of code-switching in the Irish-language. The authors found that the choice of subword model effected downstream performance of the gaBERT model \citep{barry-etal-2022-gabert}. Besides these essentially extrinsic evaluations of tokenization for Irish, intrinsic evaluation of tokenization for the language has not been investigated until now.

\subsection{Evaluation of Tokenization}
There is no clear consensus on the best method for evaluating tokenization, however, common approaches distinguish between intrinsic and extrinsic analysis. Intrinsic evaluation provides information about the quality of the tokens produced, whereas extrinsic evaluation compares different tokenization algorithms on the same downstream task. Several intrinsic metrics have been proposed, however, these metrics are not always indicators of downstream performance. 

\textbf{Rényi Efficiency/Entropy} is a generalisation of Shannon entropy that evaluates the efficiency of a tokenizer's unigram distribution by penalising overly skewed token frequency distributions (i.e. penalising low-frequency and high-frequency tokens), encouraging a more balanced vocabulary. A higher Rényi efficiency (RE) indicates a more evenly distributed vocabulary, one which reduces the risk of sparse learning from rare tokens or loss of distinction from overly frequent ones. It has been shown to correlate strongly with downstream machine translation performance as measured by BLEU \citep{zouhar-etal-2023-tokenization}; however, this metric has been criticised, with counterexamples achieving a high RE but poor downstream machine translation performance \citep{cognetta-etal-2024-two}.

\textbf{Corpus Token Count} measures the total number of tokens used to represent a corpus. A lower Corpus Token Count (CTC) is indicative of greater compression \citep{schmidt-etal-2024-tokenization}. Text compression has been used as an indicator of the quality of subwords produced by a given tokenizer in the development of the BPE algorithm \citep{sennrich-etal-2016-neural}. Moreover, it has been found that there is a correlation between compression and performance on downstream generative tasks and for smaller models of $\sim$10 million parameters \citep{goldman-etal-2024-unpacking} but \citet{schmidt-etal-2024-tokenization} found that CTC should not be used as an indicator for good downstream performance, as experimental results show that compression does not correlate with an effective tokenizer. 

\textbf{Fertility} is the average number of tokens-per-word i.e. the degree of segmentation. A fertility of 1 is considered desirable and would indicate that each word in the input text is contained in the tokenizer’s vocabulary \citep{rust-etal-2021-good}. However, a low fertility value might not be ideal for languages with complex morphology, as a higher fertility (i.e. more segmentations within a word) allows for capturing of morphemes and other re-occuring subwords \citep{brahma2025morphtokmorphologicallygroundedtokenization}. 

\textbf{Average Token Length} is the average number of characters per-token. A higher average token length has been associated with gold-standard morphological segmentation for the English language \citep{bostrom-durrett-2020-byte}. 

\subsubsection{Morphological metrics}
\textbf{MorphScore} \citep{arnett-bergen-2025-language, arnett2025evaluatingmorphologicalalignmenttokenizers} is an evaluation framework that evaluates the morphological-alignment of segmentations produced by a tokenizer with morphologically separated words, sourced from Universal Dependencies. This framework covers 86 languages, providing support for the evaluation of tokenzation of many low-resource language. Version 1 of MorphScore allowed for two possible subwords (affix and stem), while version 2 allows for three possible subwords. MorphScore has been criticised due to its lack of more-granular affix boundaries which are a feature of highly agglutinative languages, such as Turkish, where suffix-suffix boundaries are prevalent \citep{poelman-etal-2025-confounding}. This framework calculates recall and precision (including macro, micro and standard deviation), micro- and macro-F1, and mean token character ratio.

\textbf{Suffix Recall} was proposed for the Turkish language, and measures the proportion of suffixes that are correctly identified and segmented as distinct tokens \citep{erkaya-comprehensive-analysis-2022}.\textbf{Suffix Precision} was also proposed to measure the accuracy of suffix tokenization alignment by measuring the ratio of tokens that are true suffixes within a tokenizer's vocabulary, indicating how many suffix-tokens are valid morphological units rather than spurious tokens \citep{erkaya-comprehensive-analysis-2022}.

\section{MoirfEolas}
The MoirfEolas dataset (from the Irish `\textit{Moirfeolaíocht}' meaning `morphology' and `\textit{Eolas}' meaning `information') is constructed using the Irish-side of the ParaCrawl V9 \citep{banon-etal-2020-paracrawl} corpus containing 57,587 sentences with a total of 1,447,047 space-separated words, in tandem with linguistically-informed handcrafted lists of Irish eclipses, prefixes and suffixes and further prefix and suffix entries from Irish UniMorph data\footnote{\url{https://github.com/unimorph/gle}} \citep{mccarthy-etal-2020-unimorph}. UniMorph contains 197 prefixes and 73 suffixes while the handcrafted lists contain 317 prefixes, 217 suffixes and 13 eclipses. After combining the resources and removing duplicates, 330 prefixes, 213 suffixes and 13 eclipses were gathered.

A greedy-longest match method was employed to identify these eclipses, prefixes and suffixes within each unique word in the ParaCrawl corpus, resulting in the MoirfEolas dataset of over 35,000 words. The dataset contains 216, 178, and 13 unique prefixes, suffixes, and eclipses, respectively. The dataset was cleaned of English entries utilising existing morphological analysis tools for the language \citep{ui-dhonnchadha-van-genabith-2006-part}, as well as manually inspected for any incorrect mappings, resulting in the removal of over 15,000 words. Incorrect mappings generally arise from eclipses including hyphens, due to spelling variations prevalent in the data i.e. using hyphens or not using hyphens when eclipsing words. This hyphen may not always be included in-text, resulting in incorrect mappings for the eclipses `n(-)' preceding a vowel in words such as `neart' meaning \textit{strength}.  


\section{CríochScore}
\begin{figure}
    \centering
    \includegraphics[width=0.8\linewidth]{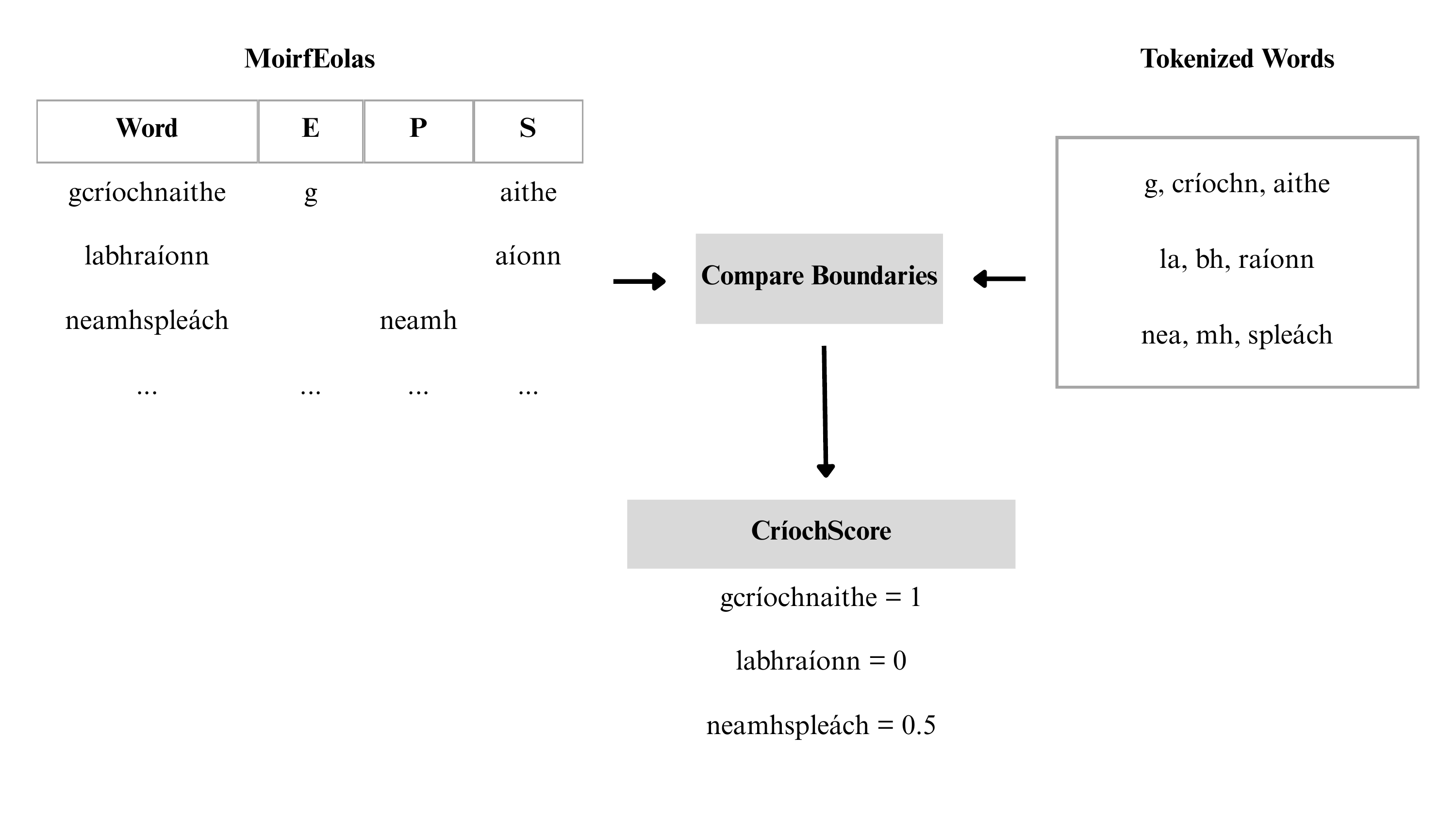}
    \caption{CríochScore calculation (E=eclipsis, P=prefix, S=suffix)}
    \label{fig:críoch}
\end{figure}

The CríochScore evaluation metric (from the Irish `\textit{Críoch}' meaning `end' or `boundary') utilises the MoirfEolas dataset to measure the level of alignment between the morphology boundaries of a word (involving eclipses, prefixes and suffixes) with the tokenization of a word, as shown in Figure \ref{fig:críoch}. The scoring criteria utilised in CríochScore are as follows:
\begin{itemize}
    \item Tokenized Correctly: A score of 1 if the prefix, suffix or eclipsis is completely segmented from the stem and not broken down further.
    \item Tokenized in Parts: A score of 0.5 if the prefix, suffix or eclipsis is completely segmented from the stem but is broken down further.
    \item Tokenized Incorrectly: A score of 0 if the prefix, suffix or eclipsis is not completely segmented from the stem.
\end{itemize}
Each word-level score is normalised by the number of evaluated morphological components, {N}, within the word (\ref{equation:criochscoreword}). The overall CríochScore for a given tokenizer is the average of all word-level CríochScores, CríochScore\textsubscript{w}.

\begin{equation}
    \frac{\sum_{i=1}^{N} \text{CríochScore}_i}{N}, \quad \text{where } N \leq 3
    \label{equation:criochscoreword}
\end{equation}

\section{Evaluation}
\label{sec:evaluation}
A total of six tokenization algorithms are evaluated using the CríochScore metric and further absorption analysis is performed to ascertain how morphemes are handled within the algorithms. HuggingFace implementations of the Unigram LM, SentencePiece Unigram LM, WordPiece, BPE, SentencePiece BPE and Byte-level BPE are evaluated across vocabulary sizes 8k, 16k, 32k and 64k\footnote{A vocabulary size of 43k was the maximum available for SentencePiece-Unigram LM for this corpus.} using the Irish side of the cleaned ParaCrawl V9 corpus \citep{banon-etal-2020-paracrawl}. This corpus was selected as the largest parallel corpus of sufficient quality, according to a survey of Irish natural language processing corpora \citep{walsh2025survey} using quality annotation labels proposed by \citet{kreutzer-etal-2022-quality}, and thus is a valuable resource for both intrinsic and extrinsic (through machine translation) evaluation of tokenization for Irish. Each subsequent tokenizer vocabulary is then evaluated using CríochScore against entries in the MoirfEolas dataset.

Furthermore, intrinsic metrics from the literature are employed to provide more context as to how the Irish language is handled by the tokenization algorithms. Corpus Token Count, Rényi Efficiency, Fertility as well as Prefix-, Suffix- and Eclipsis-level Recall and Precision are calculated by adapting previous work on Turkish to Irish. MorphScore was explored for this research, but ultimately was not used, as entries for the Irish language were limited (2576), and prefix entries labelled `preceding\textunderscore{ }part' contained a mixture of prefixes, eclipses, and lenited consonants, which would require more granular boundaries for cases of morpheme clusters (e.g. eclipses preceding prefixes)\footnote{\url{https://huggingface.co/datasets/catherinearnett/morphscore/blob/main/irish_data.csv}}. 

\section{Results and Discussion}
\subsection{CríochScore}
The CríochScore evaluation results are reported in Table \ref{table:críoch}. The Unigram LM with a vocabulary size of 8k aligned most often with Irish morphological boundaries, as evidenced by the overall CríochScore of 40.55\%. SentencePiece-BPE at 16k following closely behind with a score of 40.39\%. BPE with a vocabulary of 64k aligned least often with the morphological boundaries with a value of 9.6\%. While no algorithm achieved very high scores, this does not necessarily imply that such tokenizers do not perform well on downstream tasks. BPE aligns well with prefixes (especially for a vocabulary of 8k), however it displays some of the lowest CríochScore values across all categories evaluated; Byte- and SentencePiece-implementations of BPE align more often with Irish morphological boundaries than vanilla-BPE. 

Unigram and its SentencePiece implementation performed relatively well across all categories, showcasing the algorithm's suitability to capturing the morphology of Irish, only being outperformed by BPE at 8k in prefix handling. Unigram also handles multiple morphological components the best, where a word contains more than one of the evaluated morphological components (eclipsis, prefix and suffix). Each of the highest scores achieved used a vocabulary size of 8k, indicating that this smaller vocabulary size enables efficient token-boundary alignment with Irish morphology, segmenting eclipses, prefixes and suffixes correctly more often than tokenizers with larger vocabularies.

As the table shows, the CríochScore for words containing only an eclipsis (and no prefix or suffix) are all 50\%. On further inspection of these tokenizations, this was the result of a data oddity, where each of these words (13) contained an eclipsis paired with a hyphen (-) or apostrophe (’). These eclipses-punctuation pairs were split at the punctuation symbol, resulting in a score of 50\%, e.g. the word \textit{t-aláram} (i.e. \textit{aláram} `alarm', with an additional eclipsis `\textit{t-}’ added for grammatical reasons) is tokenized as `\textit{t}’, `-’ `\textit{aláram}’---effectively aligning with the eclipsis boundary between `-' and `\textit{a}', but also separating the leading character `\textit{t}' from `-', resulting in a CríochScore value of 0.5. Overall, prefix-boundaries are aligned with more often by each tokenizer than suffix-boundaries. Interestingly, while MoirfEolas contains more unique prefixes than suffixes, suffixes appear more frequently in the corpus indicating a potential word-initial positional bias in tokenization.   

\begin{table}[]
\resizebox{\columnwidth}{!}{%
\begin{tabular}{@{}cccccccccc@{}}
\textbf{T} &
  \textbf{V} &
  \textbf{Críoch} &
  \textbf{CríochE} &
  \textbf{CríochP} &
  \textbf{CríochS} &
  \textbf{CríochE+P} &
  \textbf{CríochE+S} &
  \textbf{CríochP+S} &
  \textbf{CríochE+P+S} \\
 &
  8 &
  \textbf{40.55} &
  {\underline{\textbf{50}}} &
  57.32 &
  38.05 &
  \textbf{35.21} &
  \textbf{47.8} &
  44.29 &
  \textbf{33.83} \\
                            & 16 & 27.64 & {\underline{\textbf{50}}} & 48.63          & 33.09          & 26.5  & 40.21 & 37.39          & 27.56 \\
                            & 32 & 27.8  & {\underline{\textbf{50}}} & 39.93          & 28.08          & 21.21 & 32.5  & 30.98          & 21.19 \\
\multirow{-4}{*}{Unigram}   & 64 & 12.71 & {\underline{\textbf{50}}} & 38.62          & 27.37          & 20.6  & 32.5  & 30.32          & 20.86 \\
\midrule \\
                            & 8  & 34.73 & {\underline{\textbf{50}}} & 56.86          & 32.86          & 6.43  & 22.7  & 42.22          & 15.05 \\
                            & 16 & 20.41 & {\underline{\textbf{50}}} & 48.05          & 28.01          & 6.43  & 19.16 & 33.84          & 12.75 \\
                            & 32 & 26.42 & {\underline{\textbf{50}}} & 39.87          & 20.3           & 6.43  & 14.99 & 24.94          & 9.89  \\
\multirow{-4}{*}{WordPiece} & 64 & 33.15 & {\underline{\textbf{50}}} & 27.69          & 11.27          & 6.43  & 10.4  & 15.82          & 7.2   \\
\midrule \\
                            & 8  & 28.98 & {\underline{\textbf{50}}} & 57.55          & \textbf{40.37} & 21.21 & 34.29 & \textbf{46.37} & 27.08 \\
                            & 16 & 12.23 & {\underline{\textbf{50}}} & 48.57          & 33.93          & 16.73 & 29.17 & 38.69          & 20.33 \\
                            & 32 & 35.81 & {\underline{\textbf{50}}} & 39.24          & 28.14          & 14    & 25.37 & 30.49          & 16.67 \\
\multirow{-4}{*}{SP-ULM}    & 43 & 26.75 & {\underline{\textbf{50}}} & 38.04          & 26.73          & 13.12 & 24.15 & 28.91          & 15.85 \\
\midrule \\
                            & 8  & 28.36 & {\underline{\textbf{50}}} & 57.21          & 35.13          & 16.81 & 27.89 & 44.92          & 20.39 \\
                            & 16 & 40.39 & {\underline{\textbf{50}}} & 50.4           & 28.24          & 11.88 & 22.49 & 37.56          & 15.71 \\
                            & 32 & 29.15 & {\underline{\textbf{50}}} & 41.88          & 19.59          & 10.3  & 16.82 & 28.6           & 11.4  \\
\multirow{-4}{*}{SP-BPE}    & 64 & 19.1  & {\underline{\textbf{50}}} & 31.58          & 10.6           & 8.19  & 11.17 & 19.24          & 7.4   \\
\midrule \\
                            & 8  & 32.95 & {\underline{\textbf{50}}} & \textbf{59.15} & 30.9           & 20.69 & 28.51 & 43.18          & 21.67 \\
                            & 16 & 33.78 & {\underline{\textbf{50}}} & 53.43          & 24.11          & 15.4  & 23.21 & 36.84          & 16.62 \\
                            & 32 & 21.08 & {\underline{\textbf{50}}} & 45.37          & 16.25          & 11.36 & 16.43 & 28.57          & 11.16 \\
\multirow{-4}{*}{BPE} &
  64 &
  {\underline{9.6}} &
  {\underline{\textbf{50}}} &
  {\underline{25.06}} &
  {\underline{8.34}} &
  {\underline{6.43}} &
  {\underline{9.16}} &
  {\underline{12.86}} &
  {\underline{5.87}} \\
  \midrule \\
                            & 8  & 35.98 & {\underline{\textbf{50}}} & 57.27          & 35.33          & 16.81 & 28.03 & 45.04          & 20.41 \\
                            & 16 & 29.36 & {\underline{\textbf{50}}} & 50.74          & 28.47          & 11.88 & 22.71 & 37.71          & 15.85 \\
                            & 32 & 21.27 & {\underline{\textbf{50}}} & 41.88          & 19.93          & 10.3  & 16.85 & 28.53          & 11.31 \\
\multirow{-4}{*}{Byte-BPE}  & 64 & 13.98 & {\underline{\textbf{50}}} & 32.72          & 12.19          & 8.71  & 12.01 & 20.05          & 7.97 
\end{tabular}%
}
\caption{Críoch results (\%) for Tokenizers (T) across vocabulary sizes (V) ranging from 8k to 64k where SP = SentencePiece, ULM = Unigram LM, E = eclipsis, P = prefix and S = suffix.}
\label{table:críoch}
\end{table}

\subsection{Intrinsic Metrics}
The other intrinsic metrics are displayed in Table \ref{table:intrinsic} and provide further context as to the behaviour of the tokenizers' handling of Irish morphology. The impact of increasing vocabulary size is again prevalent in this layer of the evaluation: corpus token count, fertility, RE, affix- and eclipses-level precision are seen to decrease as vocabulary size increases. Larger vocabulary sizes contain more whole-words as tokens, with fewer segmentations being made as indicated by the higher average token lengths and lower fertility values seen in the table. This aligns with the findings from Table \ref{table:críoch}, showing that tokenizer algorithms with larger vocabularies tend to misalign with morphological boundaries more often than smaller vocabularies. Moreover, the higher amount of whole-word storage exhibited by the larger vocabularies evaluated may harm downstream model generalisability \citep{reddy2025enoughdiminishingreturnstokenization}. 

Overall, the table clearly indicates that Unigram is segmenting words at a higher rate than the other evaluated algorithms, as seen by its higher fertility rate and lower average token length across all sizes. This finding aligns with \citet{brahma2025morphtokmorphologicallygroundedtokenization}'s hypothesis that morphologically complex languages may necessitate higher fertility to adequately capture the morphology of the language within tokenizer vocabularies and the observed CríochScore results. Unigram does, however, have the highest corpus token count values, indicating a tradeoff between compression and morphological alignment. Another trade-off emerges from the Rényi Efficiency (RE) results; Unigram has lower values than BPE varieties, with RE values decreasing as  vocabulary size increases. 

Suffix recall decreases slightly for all tokenizers as vocabulary size increases, though it remains high overall. This indicates that suffix material is generally recoverable, but not necessarily through clean suffix segmentation: recall does not penalise fragmentary tokenization, whereas precision does. The low precision values show that suffixes are typically recovered through suffix fragments rather than as clean suffix-tokens. High recall thus suggests that suffixes, or parts of them, are present in the vocabularies, but it does not imply that suffixes are segmented correctly or consistently across words. The same general pattern holds for prefixes, although recall tends to be lower and precision scores are often higher than for suffixes. Eclipses present the strongest contrast, with near-perfect recall but near-zero precision. This indicates that eclipses, which are usually only one or two characters in length, are almost always present in the vocabularies, but are almost never segmented independently from the stem or prefix. Rather than functioning as stable morpheme tokens, they appear to behave more like noise-level consonant material.

\begin{table}[]
\centering
\resizebox{\columnwidth}{!}{%
\begin{tabular}{cccccccccccc}
\textbf{Tokenizer} & \textbf{V} & \textbf{CTC} & \textbf{Fertility} & \textbf{ATL} & \textbf{RE} & \textbf{S\textsubscript{R}} & \textbf{S\textsubscript{Pr}} & \textbf{P\textsubscript{R}} & \textbf{P\textsubscript{Pr}} & \textbf{E\textsubscript{R}} & \textbf{E\textsubscript{Pr}} \\
\multirow{4}{*}{Unigram} & 8 & \underline{245.05} & \textbf{3.42} & {\underline{2.46}} & 45.6 & \textbf{91.57} & 3.17 & \textbf{81.48} & 7.13 & \textbf{100} & 1.74 \\
 & 16 & 202.32 & 2.83 & 2.98 & 42.52 & 89.33 & 2.6 & 79.17 & 5.51 & \textbf{100} & 0.91 \\
 & 32 & 170.81 & 2.39 & 3.53 & 40.1 & 84.27 & 2.04 & 75.93 & 4.5 & \textbf{100} & 0.54 \\
 & 64 & 169.71 & 2.37 & 3.55 & 39.96 & 84.83 & {\underline{2.03}} & 77.31 & {\underline{4.43}} & \textbf{100} & 0.53 \\
 \midrule
\multirow{4}{*}{WordPiece} & 8 & 222.82 & 3.11 & 2.7 & 52.86 & 91.01 & 5.47 & 71.3 & 12.43 & 84.62 & 2.16 \\
 & 16 & 191.17 & 2.67 & 3.15 & 47.59 & 88.2 & 3.9 & 65.74 & 9.23 & 76.92 & 1.26 \\
 & 32 & 158.58 & 2.22 & 3.8 & 43.51 & 88.2 & 3.13 & 65.74 & 7.66 & 53.85 & 0.87 \\
 & 64 & 120.33 & 1.68 & 5.01 & 40.46 & 84.83 & 2.64 & 61.57 & 7.2 & 46.15 & {\underline{0.39}} \\
 \midrule
\multirow{4}{*}{SP-ULM} & 8 & 239.88 & 3.35 & 2.51 & 42.55 & 90.45 & \textbf{7.32} & 76.39 & 9.39 & \textbf{100} & 1.4 \\
 & 16 & 201.06 & 2.81 & 3 & 39.45 & 87.64 & 4.49 & 77.78 & 7.4 & \textbf{100} & 0.75 \\
 & 32 & 162.78 & 2.27 & 3.7 & 37.04 & 84.83 & 3.03 & 75.46 & 5.37 & \textbf{100} & 0.49 \\
 & 43 & 155.52 & 2.17 & 3.87 & {\underline{36.44}} & 84.27 & 2.67 & 75 & 4.99 & \textbf{100} & 0.44 \\
 \midrule
\multirow{4}{*}{SP-BPE} & 8 & 225.79 & 3.15 & 2.67 & 52.71 & 89.89 & 6.79 & 72.69 & 14.99 & 92.31 & 2.56 \\
 & 16 & 195.05 & 2.72 & 3.09 & 47.55 & 89.33 & 4.55 & 70.83 & 11.76 & 92.31 & 1.5 \\
 & 32 & 161.81 & 2.26 & 3.72 & 43.52 & 87.64 & 3.28 & 68.52 & 10.1 & 92.31 & 1.09 \\
 & 64 & 122.73 & 1.71 & 4.91 & 40.66 & 82.58 & 2.79 & 62.96 & 9.72 & 76.92 & 1.05 \\
 \midrule
\multirow{4}{*}{BPE} & 8 & 210.45 & 2.94 & 2.86 & 51.99 & 89.89 & 4.21 & 71.76 & 13.3 & \textbf{100} & 2.43 \\
 & 16 & 181.54 & 2.54 & 3.32 & 47.08 & 87.64 & 3.01 & 69.91 & 11.54 & \textbf{100} & 1.53 \\
 & 32 & 150.26 & 2.1 & 4.01 & 43.23 & 87.64 & 2.41 & 68.52 & 10.37 & \textbf{100} & 1.11 \\
 & 64 & {\textbf{113.1}} & {\underline{1.58}} & \textbf{5.33} & 40.27 & {\underline{79.78}} & 2.31 & {\underline{55.09}} & 8.27 & {\underline{38.46}} & 0.45 \\
 \midrule
\multirow{4}{*}{Byte-BPE} & 8 & 227.21 & 3.17 & 2.65 & \textbf{53.05} & 89.89 & 6.91 & 72.69 & \textbf{15.01} & 92.31 & \textbf{2.61} \\
 & 16 & 195.9 & 2.74 & 3.08 & 47.86 & 89.33 & 4.67 & 70.83 & 11.63 & 92.31 & 1.52 \\
 & 32 & 162.35 & 2.27 & 3.71 & 43.77 & 87.64 & 3.32 & 68.52 & 10.01 & 92.31 & 1.09 \\
 & 64 & 132.92 & 1.86 & 4.53 & 41.38 & 85.39 & 2.86 & 65.28 & 10.43 & 84.62 & 1.05
\end{tabular}%
}
\caption{Intrinsic metric results where CTC = corpus token count (k), ATL = average token length, RE = Rényi Efficiency (\%), S = suffix, P = prefix, E = eclipsis, R = recall (\%) and Pr = precision (\%).
}
\label{table:intrinsic}
\end{table}

\subsection{Absorption Analysis}
As previously stated, eclipses are almost always absorbed/merged into a token with subsequent characters. This is to be expected due to the nature of eclipses as a high-frequency phenomenom, where a limited set of characters are regularly concatenated to a specific set of word-initial characters (see Table \ref{table:appendix-absorp} for eclipses mappings and examples). Both SentencePiece implementations as well as WordPiece 64k and BPE 64k absorb 100\% of eclipses into a token with subsequent characters. Even at smaller vocabulary sizes, all evaluated tokenizers tend to fully absorb prefixes into subsequent subwords and suffixes into preceding subwords, not distinguishing the morphological boundary represented by these affixes. This is also seen across all evaluated tokenizers as the vocabulary size increases, in line with the observation that larger vocabularies have longer subwords that misalign with the morphological boundaries of the Irish language. Unigram implementations have lower prefix and suffix absorption rates that are partially offset by higher suffix split absorption,  where a part of a suffix is separate from the stem/root of the word but the remaining part(s) is merged with preceding subwords e.g. for the word ‘labhraíonn’ with the suffix ‘aíonn’, the suffix is partially absorbed in the following tokenization: ‘labhraí’ ‘onn’, this would result in a CríochScore of 0 as the tokenization does not align with the suffix-boundary but part of the suffix is separated. As the Irish language displays a high-degree of suffixation \citep{ui-dhonnchadha-2002-two}, Unigram-based tokenizers appear to be the more appropriate choice for alignment with Irish morphology.

\section{Conclusion}
We introduce MoirfEolas and CríochScore, dedicated resources for the evaluation of tokenization boundary alignment with Irish morphology. The approach used to construct the MoirfEolas dataset offers a model for languages with similar morphological forms as the  Irish language. The CríochScore metric allows for allows for intrinsic evaluation of tokenizer alignment with morphology, building on existing metrics such as MorphScore and Suffix Recall. We evaluate commonly-used tokenizers' alignment with the morphological boundaries of the Irish language and find that Unigram implementations, particularly with smaller vocabulary sizes, produce the most aligned vocabularies, and smaller tokens than merge-based methods (BPEs, WordPiece). We also obtain practical insights for the development of Irish natural language processing applications, where there are tradoffs between morphological alignment and compression as well as vocabulary efficiency. This evaluation represents an initial investigation into tokenization of the Irish language, where downstream evaluation is necessary to verify the practical insights gathered as well as morphological-alignment's role in improving performance.


\bibliography{moirfeolas}

@inproceedings{ali-etal-2024-tokenizer,
    title = "Tokenizer Choice For {LLM} Training: Negligible or Crucial?",
    author = {Ali, Mehdi  and
      Fromm, Michael  and
      Thellmann, Klaudia  and
      Rutmann, Richard  and
      L{\"u}bbering, Max  and
      Leveling, Johannes  and
      Klug, Katrin  and
      Ebert, Jan  and
      Doll, Niclas  and
      Buschhoff, Jasper  and
      Jain, Charvi  and
      Weber, Alexander  and
      Jurkschat, Lena  and
      Abdelwahab, Hammam  and
      John, Chelsea  and
      Ortiz Suarez, Pedro  and
      Ostendorff, Malte  and
      Weinbach, Samuel  and
      Sifa, Rafet  and
      Kesselheim, Stefan  and
      Flores-Herr, Nicolas},
    editor = "Duh, Kevin  and
      Gomez, Helena  and
      Bethard, Steven",
    booktitle = "Findings of the Association for Computational Linguistics: NAACL 2024",
    month = jun,
    year = "2024",
    address = "Mexico City, Mexico",
    publisher = "Association for Computational Linguistics",
    url = "https://aclanthology.org/2024.findings-naacl.247/",
    doi = "10.18653/v1/2024.findings-naacl.247",
    pages = "3907--3924"
}

@inproceedings{sennrich-etal-2016-neural,
    title = "Neural Machine Translation of Rare Words with Subword Units",
    author = "Sennrich, Rico  and
      Haddow, Barry  and
      Birch, Alexandra",
    editor = "Erk, Katrin  and
      Smith, Noah A.",
    booktitle = "Proceedings of the 54th Annual Meeting of the Association for Computational Linguistics (Volume 1: Long Papers)",
    month = aug,
    year = "2016",
    address = "Berlin, Germany",
    publisher = "Association for Computational Linguistics",
    url = "https://aclanthology.org/P16-1162/",
    doi = "10.18653/v1/P16-1162",
    pages = "1715--1725"
}

@inproceedings{kudo-richardson-2018-sentencepiece,
    title = "{S}entence{P}iece: A simple and language independent subword tokenizer and detokenizer for Neural Text Processing",
    author = "Kudo, Taku  and
      Richardson, John",
    editor = "Blanco, Eduardo  and
      Lu, Wei",
    booktitle = "Proceedings of the 2018 Conference on Empirical Methods in Natural Language Processing: System Demonstrations",
    month = nov,
    year = "2018",
    address = "Brussels, Belgium",
    publisher = "Association for Computational Linguistics",
    url = "https://aclanthology.org/D18-2012/",
    doi = "10.18653/v1/D18-2012",
    pages = "66--71"
}

@inproceedings{kudo-2018-subword,
    title = "Subword Regularization: Improving Neural Network Translation Models with Multiple Subword Candidates",
    author = "Kudo, Taku",
    editor = "Gurevych, Iryna  and
      Miyao, Yusuke",
    booktitle = "Proceedings of the 56th Annual Meeting of the Association for Computational Linguistics (Volume 1: Long Papers)",
    month = jul,
    year = "2018",
    address = "Melbourne, Australia",
    publisher = "Association for Computational Linguistics",
    url = "https://aclanthology.org/P18-1007/",
    doi = "10.18653/v1/P18-1007",
    pages = "66--75"
}

@inproceedings{schusternakijama2012wordpiece,
  author={Schuster, Mike and Nakajima, Kaisuke},
  booktitle={2012 IEEE International Conference on Acoustics, Speech and Signal Processing (ICASSP)}, 
  title={Japanese and Korean voice search}, 
  year={2012},
  volume={},
  number={},
  pages={5149-5152},
  doi={10.1109/ICASSP.2012.6289079}}

@inproceedings{osvth-etal-2026-impact,
  title = {The Impact of Tokenization Algorithms on Hungarian Language Model Performance},
  author = {Osváth, Mátyás and Molnár, Máté Norbert and Gunics, Roland and Ligeti-Nagy, Noémi},
  booktitle = {Proceedings of the Fifteenth Language Resources and Evaluation Conference (LREC 2026)},
  month = {May},
  year = {2026},
  pages = {2545--2556},
  address = {Palma, Mallorca, Spain},
  publisher = {European Language Resources Association (ELRA)},
  editor = {Piperidis, Stelios and Bel, Núria and van den Heuvel, Henk and Ide, Nancy and Krek, Simon and Toral, Antonio},
  doi = {10.63317/29hx92kq2dxe}
}

@misc{arnett2025evaluatingmorphologicalalignmenttokenizers,
      title={Evaluating Morphological Alignment of Tokenizers in 70 Languages}, 
      author={Catherine Arnett and Marisa Hudspeth and Brendan O'Connor},
      year={2025},
      eprint={2507.06378},
      archivePrefix={arXiv},
      primaryClass={cs.CL},
      url={https://arxiv.org/abs/2507.06378} 
}

@inproceedings{arnett-bergen-2025-language,
    title = "Why do language models perform worse for morphologically complex languages?",
    author = "Arnett, Catherine  and
      Bergen, Benjamin",
    editor = "Rambow, Owen  and
      Wanner, Leo  and
      Apidianaki, Marianna  and
      Al-Khalifa, Hend  and
      Eugenio, Barbara Di  and
      Schockaert, Steven",
    booktitle = "Proceedings of the 31st International Conference on Computational Linguistics",
    month = jan,
    year = "2025",
    address = "Abu Dhabi, UAE",
    publisher = "Association for Computational Linguistics",
    url = "https://aclanthology.org/2025.coling-main.441/",
    pages = "6607--6623"
}

@inproceedings{hou-etal-2023-effects,
    title = "Effects of sub-word segmentation on performance of transformer language models",
    author = "Hou, Jue  and
      Katinskaia, Anisia  and
      Vu, Anh-Duc  and
      Yangarber, Roman",
    editor = "Bouamor, Houda  and
      Pino, Juan  and
      Bali, Kalika",
    booktitle = "Proceedings of the 2023 Conference on Empirical Methods in Natural Language Processing",
    month = dec,
    year = "2023",
    address = "Singapore",
    publisher = "Association for Computational Linguistics",
    url = "https://aclanthology.org/2023.emnlp-main.459/",
    doi = "10.18653/v1/2023.emnlp-main.459",
    pages = "7413--7425"
}

@mastersthesis{erkaya-comprehensive-analysis-2022,
    author = "Erkaya, Erencan",
    title = "A comprehensive analysis of subword tokenizers for morphologically rich languages",
    school = " Bogaziçi University",
    year = "2022"
}

@misc{vemula2025rethinkingtokenizationrichmorphology,
      title={Rethinking Tokenization for Rich Morphology: The Dominance of Unigram over BPE and Morphological Alignment}, 
      author={Saketh Reddy Vemula and Sandipan Dandapat and Dipti Misra Sharma and Parameswari Krishnamurthy},
      year={2025},
      eprint={2508.08424},
      archivePrefix={arXiv},
      primaryClass={cs.CL},
      url={https://arxiv.org/abs/2508.08424} 
}

@inproceedings{petrov-unfairness-2023,
author = {Petrov, Aleksandar and Malfa, Emanuele La and Torr, Philip H.S. and Bibi, Adel},
title = {Language model tokenizers introduce unfairness between languages},
year = {2023},
publisher = {Curran Associates Inc.},
address = {Red Hook, NY, USA},
booktitle = {Proceedings of the 37th International Conference on Neural Information Processing Systems},
articleno = {1608},
numpages = {28},
location = {New Orleans, LA, USA},
series = {NIPS '23}
}

@inproceedings{arnett-etal-2024-bit,
    title = "A Bit of a Problem: Measurement Disparities in Dataset Sizes across Languages",
    author = "Arnett, Catherine  and
      Chang, Tyler A.  and
      Bergen, Benjamin",
    editor = "Melero, Maite  and
      Sakti, Sakriani  and
      Soria, Claudia",
    booktitle = "Proceedings of the 3rd Annual Meeting of the Special Interest Group on Under-resourced Languages LREC-COLING 2024",
    month = may,
    year = "2024",
    address = "Torino, Italia",
    publisher = "ELRA and ICCL",
    url = "https://aclanthology.org/2024.sigul-1.1/",
    pages = "1--9"
}

@inproceedings{chai-etal-2024-tokenization,
    title = "Tokenization Falling Short: On Subword Robustness in Large Language Models",
    author = "Chai, Yekun  and
      Fang, Yewei  and
      Peng, Qiwei  and
      Li, Xuhong",
    editor = "Al-Onaizan, Yaser  and
      Bansal, Mohit  and
      Chen, Yun-Nung",
    booktitle = "Findings of the Association for Computational Linguistics: EMNLP 2024",
    month = nov,
    year = "2024",
    address = "Miami, Florida, USA",
    publisher = "Association for Computational Linguistics",
    url = "https://aclanthology.org/2024.findings-emnlp.86/",
    doi = "10.18653/v1/2024.findings-emnlp.86",
    pages = "1582--1599"
}

@inproceedings{ui-dhonnchadha-2002-two,
    title = "A Two-level Morphological Analyser and Generator for {I}rish using Finite-State Transducers",
    author = "U{\'i} Dhonnchadha, Elaine",
    editor = "Gonz{\'a}lez Rodr{\'i}guez, Manuel  and
      Suarez Araujo, Carmen Paz",
    booktitle = "Proceedings of the Third International Conference on Language Resources and Evaluation ({LREC}{'}02)",
    month = may,
    year = "2002",
    address = "Las Palmas, Canary Islands - Spain",
    publisher = "European Language Resources Association (ELRA)",
    url = "https://aclanthology.org/L02-1147/"
}

@misc{digitalplan,
    author = {Ní Chasaide, Ailbhe and Ní
Chiarán, Neasa and Uí Dhonnchadha, Elaine and Lynn, Teresa and Judge, John},
    title = {{Digital Plan for the Irish Language Speech and Language Technologies 2023-2027}},
    howpublished = {Available at \url{https://assets.gov.ie/241755/e82c256a-6f47-4ddb-8ce6-ff81df208bb1.pdf}},
    year={2022}
}

@misc{Radford2019LanguageMA,
  title={Language Models are Unsupervised Multitask Learners},
  author={Alec Radford and Jeff Wu and Rewon Child and David Luan and Dario Amodei and Ilya Sutskever},
  year={2019},
  url={https://api.semanticscholar.org/CorpusID:160025533}
}

@misc{liu2019robertarobustlyoptimizedbert,
      title={RoBERTa: A Robustly Optimized BERT Pretraining Approach}, 
      author={Yinhan Liu and Myle Ott and Naman Goyal and Jingfei Du and Mandar Joshi and Danqi Chen and Omer Levy and Mike Lewis and Luke Zettlemoyer and Veselin Stoyanov},
      year={2019},
      eprint={1907.11692},
      archivePrefix={arXiv},
      primaryClass={cs.CL},
      url={https://arxiv.org/abs/1907.11692} 
}

@misc{lewis2019bartdenoisingsequencetosequencepretraining,
      title={BART: Denoising Sequence-to-Sequence Pre-training for Natural Language Generation, Translation, and Comprehension}, 
      author={Mike Lewis and Yinhan Liu and Naman Goyal and Marjan Ghazvininejad and Abdelrahman Mohamed and Omer Levy and Ves Stoyanov and Luke Zettlemoyer},
      year={2019},
      eprint={1910.13461},
      archivePrefix={arXiv},
      primaryClass={cs.CL},
      url={https://arxiv.org/abs/1910.13461} 
}

@misc{thoppilan2022lamdalanguagemodelsdialog,
      title={LaMDA: Language Models for Dialog Applications}, 
      author={Romal Thoppilan and Daniel De Freitas and Jamie Hall and Noam Shazeer and Apoorv Kulshreshtha and Heng-Tze Cheng and Alicia Jin and Taylor Bos and Leslie Baker and Yu Du and YaGuang Li and Hongrae Lee and Huaixiu Steven Zheng and Amin Ghafouri and Marcelo Menegali and Yanping Huang and Maxim Krikun and Dmitry Lepikhin and James Qin and Dehao Chen and Yuanzhong Xu and Zhifeng Chen and Adam Roberts and Maarten Bosma and Vincent Zhao and Yanqi Zhou and Chung-Ching Chang and Igor Krivokon and Will Rusch and Marc Pickett and Pranesh Srinivasan and Laichee Man and Kathleen Meier-Hellstern and Meredith Ringel Morris and Tulsee Doshi and Renelito Delos Santos and Toju Duke and Johnny Soraker and Ben Zevenbergen and Vinodkumar Prabhakaran and Mark Diaz and Ben Hutchinson and Kristen Olson and Alejandra Molina and Erin Hoffman-John and Josh Lee and Lora Aroyo and Ravi Rajakumar and Alena Butryna and Matthew Lamm and Viktoriya Kuzmina and Joe Fenton and Aaron Cohen and Rachel Bernstein and Ray Kurzweil and Blaise Aguera-Arcas and Claire Cui and Marian Croak and Ed Chi and Quoc Le},
      year={2022},
      eprint={2201.08239},
      archivePrefix={arXiv},
      primaryClass={cs.CL},
      url={https://arxiv.org/abs/2201.08239} 
}

@misc{gemmateam2024gemmaopenmodelsbased,
      title={Gemma: Open Models Based on Gemini Research and Technology}, 
      author={Gemma Team and Thomas Mesnard and Cassidy Hardin and Robert Dadashi and Surya Bhupatiraju and Shreya Pathak and Laurent Sifre and Morgane Rivière and Mihir Sanjay Kale and Juliette Love and Pouya Tafti and Léonard Hussenot and Pier Giuseppe Sessa and Aakanksha Chowdhery and Adam Roberts and Aditya Barua and Alex Botev and Alex Castro-Ros and Ambrose Slone and Amélie Héliou and Andrea Tacchetti and Anna Bulanova and Antonia Paterson and Beth Tsai and Bobak Shahriari and Charline Le Lan and Christopher A. Choquette-Choo and Clément Crepy and Daniel Cer and Daphne Ippolito and David Reid and Elena Buchatskaya and Eric Ni and Eric Noland and Geng Yan and George Tucker and George-Christian Muraru and Grigory Rozhdestvenskiy and Henryk Michalewski and Ian Tenney and Ivan Grishchenko and Jacob Austin and James Keeling and Jane Labanowski and Jean-Baptiste Lespiau and Jeff Stanway and Jenny Brennan and Jeremy Chen and Johan Ferret and Justin Chiu and Justin Mao-Jones and Katherine Lee and Kathy Yu and Katie Millican and Lars Lowe Sjoesund and Lisa Lee and Lucas Dixon and Machel Reid and Maciej Mikuła and Mateo Wirth and Michael Sharman and Nikolai Chinaev and Nithum Thain and Olivier Bachem and Oscar Chang and Oscar Wahltinez and Paige Bailey and Paul Michel and Petko Yotov and Rahma Chaabouni and Ramona Comanescu and Reena Jana and Rohan Anil and Ross McIlroy and Ruibo Liu and Ryan Mullins and Samuel L Smith and Sebastian Borgeaud and Sertan Girgin and Sholto Douglas and Shree Pandya and Siamak Shakeri and Soham De and Ted Klimenko and Tom Hennigan and Vlad Feinberg and Wojciech Stokowiec and Yu-hui Chen and Zafarali Ahmed and Zhitao Gong and Tris Warkentin and Ludovic Peran and Minh Giang and Clément Farabet and Oriol Vinyals and Jeff Dean and Koray Kavukcuoglu and Demis Hassabis and Zoubin Ghahramani and Douglas Eck and Joelle Barral and Fernando Pereira and Eli Collins and Armand Joulin and Noah Fiedel and Evan Senter and Alek Andreev and Kathleen Kenealy},
      year={2024},
      eprint={2403.08295},
      archivePrefix={arXiv},
      primaryClass={cs.CL},
      url={https://arxiv.org/abs/2403.08295} 
}

@misc{touvron2023llamaopenefficientfoundation,
      title={LLaMA: Open and Efficient Foundation Language Models}, 
      author={Hugo Touvron and Thibaut Lavril and Gautier Izacard and Xavier Martinet and Marie-Anne Lachaux and Timothée Lacroix and Baptiste Rozière and Naman Goyal and Eric Hambro and Faisal Azhar and Aurelien Rodriguez and Armand Joulin and Edouard Grave and Guillaume Lample},
      year={2023},
      eprint={2302.13971},
      archivePrefix={arXiv},
      primaryClass={cs.CL},
      url={https://arxiv.org/abs/2302.13971} 
}

@misc{yang2020xlnetgeneralizedautoregressivepretraining,
      title={XLNet: Generalized Autoregressive Pretraining for Language Understanding}, 
      author={Zhilin Yang and Zihang Dai and Yiming Yang and Jaime Carbonell and Ruslan Salakhutdinov and Quoc V. Le},
      year={2020},
      eprint={1906.08237},
      archivePrefix={arXiv},
      primaryClass={cs.CL},
      url={https://arxiv.org/abs/1906.08237} 
}

@misc{lan2020albertlitebertselfsupervised,
      title={ALBERT: A Lite BERT for Self-supervised Learning of Language Representations}, 
      author={Zhenzhong Lan and Mingda Chen and Sebastian Goodman and Kevin Gimpel and Piyush Sharma and Radu Soricut},
      year={2020},
      eprint={1909.11942},
      archivePrefix={arXiv},
      primaryClass={cs.CL},
      url={https://arxiv.org/abs/1909.11942} 
}

@misc{raffel2023exploringlimitstransferlearning,
      title={Exploring the Limits of Transfer Learning with a Unified Text-to-Text Transformer}, 
      author={Colin Raffel and Noam Shazeer and Adam Roberts and Katherine Lee and Sharan Narang and Michael Matena and Yanqi Zhou and Wei Li and Peter J. Liu},
      year={2023},
      eprint={1910.10683},
      archivePrefix={arXiv},
      primaryClass={cs.LG},
      url={https://arxiv.org/abs/1910.10683} 
}

@inproceedings{devlin-etal-2019-bert,
    title = "{BERT}: Pre-training of Deep Bidirectional Transformers for Language Understanding",
    author = "Devlin, Jacob  and
      Chang, Ming-Wei  and
      Lee, Kenton  and
      Toutanova, Kristina",
    editor = "Burstein, Jill  and
      Doran, Christy  and
      Solorio, Thamar",
    booktitle = "Proceedings of the 2019 Conference of the North {A}merican Chapter of the Association for Computational Linguistics: Human Language Technologies, Volume 1 (Long and Short Papers)",
    month = jun,
    year = "2019",
    address = "Minneapolis, Minnesota",
    publisher = "Association for Computational Linguistics",
    url = "https://aclanthology.org/N19-1423/",
    doi = "10.18653/v1/N19-1423",
    pages = "4171--4186"
}

@misc{clark2020electrapretrainingtextencoders,
      title={ELECTRA: Pre-training Text Encoders as Discriminators Rather Than Generators}, 
      author={Kevin Clark and Minh-Thang Luong and Quoc V. Le and Christopher D. Manning},
      year={2020},
      eprint={2003.10555},
      archivePrefix={arXiv},
      primaryClass={cs.CL},
      url={https://arxiv.org/abs/2003.10555} 
}

@article{Kilgarriff2006,
  author    = {Adam Kilgarriff and Michael Rundell and Elaine U{\'\i} Dhonnchadha},
  title     = {Efficient corpus development for lexicography: building the New Corpus for Ireland},
  journal   = {Language Resources and Evaluation},
  year      = {2006},
  volume    = {40},
  number    = {2},
  pages     = {127--152},
  issn      = {1572-0218},
  doi       = {10.1007/s10579-006-9011-7},
  url       = {https://doi.org/10.1007/s10579-006-9011-7}
}

@phdthesis{Cassidy2023,
  author = {Lauren Cassidy},
  title  = {Linguistic Analysis and Automatic Dependency Parsing of Tweets in Modern Irish},
  school = {Dublin City University},
  year   = {2023},
  type   = {PhD thesis},
  url    = {https://doras.dcu.ie/29326/1/Lauren_Cassidy_PhD_thesis.pdf}
}

@inproceedings{doyle-etal-2019-character,
    title = {A Character-Level {LSTM} Network Model for Tokenizing the {O}ld {I}rish text of the {W}{\"u}rzburg Glosses on the {P}auline Epistles},
    author = "Doyle, Adrian  and
      McCrae, John P.  and
      Downey, Clodagh",
    editor = "Lynn, Teresa  and
      Prys, Delyth  and
      Batchelor, Colin  and
      Tyers, Francis",
    booktitle = "Proceedings of the Celtic Language Technology Workshop",
    month = aug,
    year = "2019",
    address = "Dublin, Ireland",
    publisher = "European Association for Machine Translation",
    url = "https://aclanthology.org/W19-6910/",
    pages = "70--79"
}

@inproceedings{doyle-mccrae-2025-assessment,
    title = "An Assessment of Word Separation Practices in {O}ld {I}rish Text Resources and a Universal Method for Tokenising {O}ld {I}rish Text",
    author = "Doyle, Adrian  and
      McCrae, John P.",
    editor = "Davis, Brian  and
      Fransen, Theodorus  and
      Dhonnchadha, Elaine U{\'i}  and
      Walsh, Abigail",
    booktitle = "Proceedings of the 5th Celtic Language Technology Workshop",
    month = jan,
    year = "2025",
    address = "Abu Dhabi [Virtual Workshop]",
    publisher = "International Committee on Computational Linguistics",
    url = "https://aclanthology.org/2025.cltw-1.1/",
    pages = "1--11"
}

@inproceedings{lankford-etal-2021-transformers,
    title = "Transformers for Low-Resource Languages: Is F{\'e}idir Linn!",
    author = "Lankford, Seamus and
      Alfi, Haithem and
      Way, Andy",
    editor = "Duh, Kevin  and
      Guzm{\'a}n, Francisco",
    booktitle = "Proceedings of Machine Translation Summit XVIII: Research Track",
    month = aug,
    year = "2021",
    address = "Virtual",
    publisher = "Association for Machine Translation in the Americas",
    url = "https://aclanthology.org/2021.mtsummit-research.5/",
    pages = "48--60"
}

@inproceedings{barry-etal-2022-gabert,
    title = "ga{BERT} {---} an {I}rish Language Model",
    author = "Barry, James  and
      Wagner, Joachim  and
      Cassidy, Lauren  and
      Cowap, Alan  and
      Lynn, Teresa  and
      Walsh, Abigail  and
      {\'O} Meachair, M{\'i}che{\'a}l J.  and
      Foster, Jennifer",
    editor = "Calzolari, Nicoletta  and
      B{\'e}chet, Fr{\'e}d{\'e}ric  and
      Blache, Philippe  and
      Choukri, Khalid  and
      Cieri, Christopher  and
      Declerck, Thierry  and
      Goggi, Sara  and
      Isahara, Hitoshi  and
      Maegaard, Bente  and
      Mariani, Joseph  and
      Mazo, H{\'e}l{\`e}ne  and
      Odijk, Jan  and
      Piperidis, Stelios",
    booktitle = "Proceedings of the Thirteenth Language Resources and Evaluation Conference",
    month = jun,
    year = "2022",
    address = "Marseille, France",
    publisher = "European Language Resources Association",
    url = "https://aclanthology.org/2022.lrec-1.511/",
    pages = "4774--4788"
}

@inproceedings{zouhar-etal-2023-tokenization,
    title = "Tokenization and the Noiseless Channel",
    author = "Zouhar, Vil{\'e}m  and
      Meister, Clara  and
      Gastaldi, Juan  and
      Du, Li  and
      Sachan, Mrinmaya  and
      Cotterell, Ryan",
    editor = "Rogers, Anna  and
      Boyd-Graber, Jordan  and
      Okazaki, Naoaki",
    booktitle = "Proceedings of the 61st Annual Meeting of the Association for Computational Linguistics (Volume 1: Long Papers)",
    month = jul,
    year = "2023",
    address = "Toronto, Canada",
    publisher = "Association for Computational Linguistics",
    url = "https://aclanthology.org/2023.acl-long.284/",
    doi = "10.18653/v1/2023.acl-long.284",
    pages = "5184--5207"
}

@inproceedings{schmidt-etal-2024-tokenization,
    title = "Tokenization Is More Than Compression",
    author = "Schmidt, Craig W  and
      Reddy, Varshini  and
      Zhang, Haoran  and
      Alameddine, Alec  and
      Uzan, Omri  and
      Pinter, Yuval  and
      Tanner, Chris",
    editor = "Al-Onaizan, Yaser  and
      Bansal, Mohit  and
      Chen, Yun-Nung",
    booktitle = "Proceedings of the 2024 Conference on Empirical Methods in Natural Language Processing",
    month = nov,
    year = "2024",
    address = "Miami, Florida, USA",
    publisher = "Association for Computational Linguistics",
    url = "https://aclanthology.org/2024.emnlp-main.40/",
    doi = "10.18653/v1/2024.emnlp-main.40",
    pages = "678--702"
}

@inproceedings{cognetta-etal-2024-two,
    title = "Two Counterexamples to Tokenization and the Noiseless Channel",
    author = "Cognetta, Marco  and
      Zouhar, Vil{\'e}m  and
      Moon, Sangwhan  and
      Okazaki, Naoaki",
    editor = "Calzolari, Nicoletta  and
      Kan, Min-Yen  and
      Hoste, Veronique  and
      Lenci, Alessandro  and
      Sakti, Sakriani  and
      Xue, Nianwen",
    booktitle = "Proceedings of the 2024 Joint International Conference on Computational Linguistics, Language Resources and Evaluation (LREC-COLING 2024)",
    month = may,
    year = "2024",
    address = "Torino, Italia",
    publisher = "ELRA and ICCL",
    url = "https://aclanthology.org/2024.lrec-main.1469/",
    pages = "16897--16906"
}

@inproceedings{goldman-etal-2024-unpacking,
    title = "Unpacking Tokenization: Evaluating Text Compression and its Correlation with Model Performance",
    author = "Goldman, Omer  and
      Caciularu, Avi  and
      Eyal, Matan  and
      Cao, Kris  and
      Szpektor, Idan  and
      Tsarfaty, Reut",
    editor = "Ku, Lun-Wei  and
      Martins, Andre  and
      Srikumar, Vivek",
    booktitle = "Findings of the Association for Computational Linguistics: ACL 2024",
    month = aug,
    year = "2024",
    address = "Bangkok, Thailand",
    publisher = "Association for Computational Linguistics",
    url = "https://aclanthology.org/2024.findings-acl.134/",
    doi = "10.18653/v1/2024.findings-acl.134",
    pages = "2274--2286"
}

@inproceedings{rust-etal-2021-good,
    title = "How Good is Your Tokenizer? On the Monolingual Performance of Multilingual Language Models",
    author = "Rust, Phillip  and
      Pfeiffer, Jonas  and
      Vuli{\'c}, Ivan  and
      Ruder, Sebastian  and
      Gurevych, Iryna",
    editor = "Zong, Chengqing  and
      Xia, Fei  and
      Li, Wenjie  and
      Navigli, Roberto",
    booktitle = "Proceedings of the 59th Annual Meeting of the Association for Computational Linguistics and the 11th International Joint Conference on Natural Language Processing (Volume 1: Long Papers)",
    month = aug,
    year = "2021",
    address = "Online",
    publisher = "Association for Computational Linguistics",
    url = "https://aclanthology.org/2021.acl-long.243/",
    doi = "10.18653/v1/2021.acl-long.243",
    pages = "3118--3135"
}

@misc{brahma2025morphtokmorphologicallygroundedtokenization,
      title={{MorphTok: Morphologically Grounded Tokenization for Indian Languages}}, 
      author={Maharaj Brahma and N J Karthika and Atul Singh and Devaraj Adiga and Smruti Bhate and Ganesh Ramakrishnan and Rohit Saluja and Maunendra Sankar Desarkar},
      year={2025},
      eprint={2504.10335},
      archivePrefix={arXiv},
      primaryClass={cs.CL},
      url={https://arxiv.org/abs/2504.10335} 
}

@inproceedings{bostrom-durrett-2020-byte,
    title = "Byte Pair Encoding is Suboptimal for Language Model Pretraining",
    author = "Bostrom, Kaj  and
      Durrett, Greg",
    editor = "Cohn, Trevor  and
      He, Yulan  and
      Liu, Yang",
    booktitle = "Findings of the Association for Computational Linguistics: EMNLP 2020",
    month = nov,
    year = "2020",
    address = "Online",
    publisher = "Association for Computational Linguistics",
    url = "https://aclanthology.org/2020.findings-emnlp.414/",
    doi = "10.18653/v1/2020.findings-emnlp.414",
    pages = "4617--4624"
}

@inproceedings{poelman-etal-2025-confounding,
    title = "Confounding Factors in Relating Model Performance to Morphology",
    author = "Poelman, Wessel  and
      Bauwens, Thomas  and
      de Lhoneux, Miryam",
    editor = "Christodoulopoulos, Christos  and
      Chakraborty, Tanmoy  and
      Rose, Carolyn  and
      Peng, Violet",
    booktitle = "Proceedings of the 2025 Conference on Empirical Methods in Natural Language Processing",
    month = nov,
    year = "2025",
    address = "Suzhou, China",
    publisher = "Association for Computational Linguistics",
    url = "https://aclanthology.org/2025.emnlp-main.369/",
    doi = "10.18653/v1/2025.emnlp-main.369",
    pages = "7273--7298",
    ISBN = "979-8-89176-332-6"
}

@inproceedings{banon-etal-2020-paracrawl,
    title = "{P}ara{C}rawl: Web-Scale Acquisition of Parallel Corpora",
    author = "Ba{\~n}{\'o}n, Marta  and
      Chen, Pinzhen  and
      Haddow, Barry  and
      Heafield, Kenneth  and
      Hoang, Hieu  and
      Espl{\`a}-Gomis, Miquel  and
      Forcada, Mikel L.  and
      Kamran, Amir  and
      Kirefu, Faheem  and
      Koehn, Philipp  and
      Ortiz Rojas, Sergio  and
      Pla Sempere, Leopoldo  and
      Ram{\'i}rez-S{\'a}nchez, Gema  and
      Sarr{\'i}as, Elsa  and
      Strelec, Marek  and
      Thompson, Brian  and
      Waites, William  and
      Wiggins, Dion  and
      Zaragoza, Jaume",
    editor = "Jurafsky, Dan  and
      Chai, Joyce  and
      Schluter, Natalie  and
      Tetreault, Joel",
    booktitle = "Proceedings of the 58th Annual Meeting of the Association for Computational Linguistics",
    month = jul,
    year = "2020",
    address = "Online",
    publisher = "Association for Computational Linguistics",
    url = "https://aclanthology.org/2020.acl-main.417/",
    doi = "10.18653/v1/2020.acl-main.417",
    pages = "4555--4567"
}

@inproceedings{walsh2025survey,
  title     = {Survey of Irish Text Corpora for NLP},
  author    = {Walsh, Abigail and Andrade, Mark and O'Connell, Ornait and O'Connor, {\'E}anna and Adkins, Jane},
  booktitle = {Proceedings of the 1st Workshop on Multilingual Data Quality Signals (WMDQS) at COLM 2025},
  year      = {2025},
  publisher = {Workshop on Multilingual Data Quality Signals},
  url        = {https://wmdqs.org/submissions-2025/9.pdf}
}

@misc{taylor2022galacticalargelanguagemodel,
      title={Galactica: A Large Language Model for Science}, 
      author={Ross Taylor and Marcin Kardas and Guillem Cucurull and Thomas Scialom and Anthony Hartshorn and Elvis Saravia and Andrew Poulton and Viktor Kerkez and Robert Stojnic},
      year={2022},
      eprint={2211.09085},
      archivePrefix={arXiv},
      primaryClass={cs.CL},
      url={https://arxiv.org/abs/2211.09085} 
}

@inproceedings{ui-dhonnchadha-van-genabith-2006-part,
    title = "A Part-of-speech tagger for {I}rish using Finite-State Morphology and Constraint Grammar Disambiguation",
    author = "U{\'i} Dhonnchadha, Elaine  and Van Genabith, Josef",
    editor = "Calzolari, Nicoletta  and
      Choukri, Khalid  and
      Gangemi, Aldo  and
      Maegaard, Bente  and
      Mariani, Joseph  and
      Odijk, Jan  and
      Tapias, Daniel",
    booktitle = "Proceedings of the Fifth International Conference on Language Resources and Evaluation ({LREC}{'}06)",
    month = may,
    year = "2006",
    address = "Genoa, Italy",
    publisher = "European Language Resources Association (ELRA)",
    url = "https://aclanthology.org/L06-1103/"
}

@article{kreutzer-etal-2022-quality,
    title = "Quality at a Glance: An Audit of Web-Crawled Multilingual Datasets",
    author = {Kreutzer, Julia  and
      Caswell, Isaac  and
      Wang, Lisa  and
      Wahab, Ahsan  and
      van Esch, Daan  and
      Ulzii-Orshikh, Nasanbayar  and
      Tapo, Allahsera  and
      Subramani, Nishant  and
      Sokolov, Artem  and
      Sikasote, Claytone  and
      Setyawan, Monang  and
      Sarin, Supheakmungkol  and
      Samb, Sokhar  and
      Sagot, Beno{\^i}t  and
      Rivera, Clara  and
      Rios, Annette  and
      Papadimitriou, Isabel  and
      Osei, Salomey  and
      Suarez, Pedro Ortiz  and
      Orife, Iroro  and
      Ogueji, Kelechi  and
      Rubungo, Andre Niyongabo  and
      Nguyen, Toan Q.  and
      M{\"u}ller, Mathias  and
      M{\"u}ller, Andr{\'e}  and
      Muhammad, Shamsuddeen Hassan  and
      Muhammad, Nanda  and
      Mnyakeni, Ayanda  and
      Mirzakhalov, Jamshidbek  and
      Matangira, Tapiwanashe  and
      Leong, Colin  and
      Lawson, Nze  and
      Kudugunta, Sneha  and
      Jernite, Yacine  and
      Jenny, Mathias  and
      Firat, Orhan  and
      Dossou, Bonaventure F. P.  and
      Dlamini, Sakhile  and
      de Silva, Nisansa  and
      {\c{C}}abuk Ball{\i}, Sakine  and
      Biderman, Stella  and
      Battisti, Alessia  and
      Baruwa, Ahmed  and
      Bapna, Ankur  and
      Baljekar, Pallavi  and
      Azime, Israel Abebe  and
      Awokoya, Ayodele  and
      Ataman, Duygu  and
      Ahia, Orevaoghene  and
      Ahia, Oghenefego  and
      Agrawal, Sweta  and
      Adeyemi, Mofetoluwa},
    editor = "Roark, Brian  and
      Nenkova, Ani",
    journal = "Transactions of the Association for Computational Linguistics",
    volume = "10",
    year = "2022",
    address = "Cambridge, MA",
    publisher = "MIT Press",
    url = "https://aclanthology.org/2022.tacl-1.4/",
    doi = "10.1162/tacl_a_00447",
    pages = "50--72"
}

@misc{reddy2025enoughdiminishingreturnstokenization,
      title={How Much is Enough? The Diminishing Returns of Tokenization Training Data}, 
      author={Varshini Reddy and Craig W. Schmidt and Yuval Pinter and Chris Tanner},
      year={2025},
      eprint={2502.20273},
      archivePrefix={arXiv},
      primaryClass={cs.CL},
      url={https://arxiv.org/abs/2502.20273} 
}

@inproceedings{mccarthy-etal-2020-unimorph,
    title = "{U}ni{M}orph 3.0: {U}niversal {M}orphology",
    author = "McCarthy, Arya D.  and
      Kirov, Christo  and
      Grella, Matteo  and
      Nidhi, Amrit  and
      Xia, Patrick  and
      Gorman, Kyle  and
      Vylomova, Ekaterina  and
      Mielke, Sabrina J.  and
      Nicolai, Garrett  and
      Silfverberg, Miikka  and
      Arkhangelskiy, Timofey  and
      Krizhanovsky, Nataly  and
      Krizhanovsky, Andrew  and
      Klyachko, Elena  and
      Sorokin, Alexey  and
      Mansfield, John  and
      Ern{\v{s}}treits, Valts  and
      Pinter, Yuval  and
      Jacobs, Cassandra L.  and
      Cotterell, Ryan  and
      Hulden, Mans  and
      Yarowsky, David",
    editor = "Calzolari, Nicoletta  and
      B{\'e}chet, Fr{\'e}d{\'e}ric  and
      Blache, Philippe  and
      Choukri, Khalid  and
      Cieri, Christopher  and
      Declerck, Thierry  and
      Goggi, Sara  and
      Isahara, Hitoshi  and
      Maegaard, Bente  and
      Mariani, Joseph  and
      Mazo, H{\'e}l{\`e}ne  and
      Moreno, Asuncion  and
      Odijk, Jan  and
      Piperidis, Stelios",
    booktitle = "Proceedings of the Twelfth Language Resources and Evaluation Conference",
    month = may,
    year = "2020",
    address = "Marseille, France",
    publisher = "European Language Resources Association",
    url = "https://aclanthology.org/2020.lrec-1.483/",
    pages = "3922--3931",
    language = "eng",
    ISBN = "979-10-95546-34-4"
}
\bibliographystyle{colm2026_conference}

\appendix
\section{Appendix}
\label{sec:appendix}

\subsection{Eclipses Rules in the Irish Language}
\label{app:eclipses}
Each eclipses in the Irish language appears before specific characters in the language. The eclipsis process adds a consonant to the stem of a word known as an `urú’, according to phonological rules. This process is used in specific cases such as after certain prepositions, numbers, possessive determiners etc. Each eclipsis, its subsequent characters and example usage in the language are displayed in Table \ref{table-1-eclipses}. 

\begin{table}[]
\centering
\resizebox{\columnwidth}{!}{%
\begin{tabular}{@{}cccc@{}}
\textbf{Eclipsis} & \textbf{Other Forms} & \textbf{Subsequent Character(s)} & \textbf{Example Usage} \\
\toprule
b & - & P, p & \begin{tabular}[c]{@{}c@{}}i \textbf{bp}áirt\\ ‘in part’\end{tabular} \\
\midrule
bh & - & F, f & \begin{tabular}[c]{@{}c@{}}go \textbf{bh}fuil\\ ‘that’\end{tabular} \\
\midrule
d & d’ & \begin{tabular}[c]{@{}c@{}}Fh, fh\\ T, t\end{tabular} & \begin{tabular}[c]{@{}c@{}}\textbf{D’fh}reastal mé ar…\\ ‘I attended…’\end{tabular} \\
\midrule
g & - & C, c & \begin{tabular}[c]{@{}c@{}}…i \textbf{gc}omparáid le…\\ ‘...in comparison with…’\end{tabular} \\
\midrule
h & h- & \begin{tabular}[c]{@{}c@{}}a, e, i, o, u\\ A, E, I, O, u\\ á, é, í, ó, ú\\ Á, É, Í, Ó, Ú\end{tabular} & \begin{tabular}[c]{@{}c@{}}go \textbf{hu}ile ‘s go \textbf{hi}omlán\\ ‘entirely’\end{tabular} \\
\midrule
m & - & B, b & \begin{tabular}[c]{@{}c@{}}…i \textbf{mB}áile Atha Cliath.\\ ‘...in Dublin City.’\end{tabular} \\
\midrule
n & n- & \begin{tabular}[c]{@{}c@{}}a, e, i, o, u\\ A, E, I, O, u\\ á, é, í, ó, ú\\ Á, É, Í, Ó, Ú\\ D, d\\ G, g\end{tabular} & \begin{tabular}[c]{@{}c@{}}seacht \textbf{n-o}ileán\\ ‘seven islands’\end{tabular} \\
\midrule
t & t- & \begin{tabular}[c]{@{}c@{}}a, e, i, o, u\\ A, E, I, O, u\\ á, é, í, ó, ú\\ Á, É, Í, Ó, Ú\\ S, s\end{tabular} & \begin{tabular}[c]{@{}c@{}}An \textbf{ts}ráid mhór\\ ‘The main street’\end{tabular}
\end{tabular}%
}
\caption{Eclipsis compositions and examples in the Irish Language.}
\label{table-1-eclipses}
\end{table}

\subsection{Absorption Analysis Results}
The results of the absorption analysis carried out are displayed in Table \ref{table:appendix-absorp}. Absorption represents when an eclipsis, prefix or suffix is merged into a token with a words preceding or subsequent characters e.g. for the prefix `neamh' in the word `neamhspleách' meaning \textit{independent}, the prefix is said to be absorbed if the tokenization of the word is `neamhsp', 'leách' or 'neamhsple', 'ách' etc. Split absorption represents when an eclipsis, prefix or suffix is segmented and part of the component is absorbed/merged into a token with a words preceding or subseqent characters e.g. for the suffix `aíonn' in the word `labhraíonn', the suffix is said to be split absorbed if the word is tokenized as `labhra', 'íonn' or `labhraío', `nn' etc. Eclipsis mapping is also reported in Table \ref{table:appendix-absorp}, which represents when an eclipsis is tokenized according to the mapping rules displayed in Table \ref{table-1-eclipses} e.g. for the eclipsis `g' in the word `gcríochnaithe', if the tokenization of this word begins with the token 'gc'.

\begin{table}[]
\centering
\resizebox{\columnwidth}{!}{%
\begin{tabular}{cccccccc}
\textbf{Tokenizer} & \textbf{V} & \textbf{E Absorbed} & \textbf{E Mapping} & \textbf{S Absorbed} & \textbf{S Split Absorbed} & \textbf{P Absorbed} & \textbf{P Split Absorbed} \\
\multirow{4}{*}{Unigram} & 8 & {\underline{50.45}} & \textbf{4.23} & {\underline{36.86}} & 22.54 & 40.55 & 4.13 \\
 & 16 & 60.97 & 2.53 & 40.42 & 24.96 & 51.95 & 3.83 \\
 & 32 & 72.9 & 1.61 & 44.61 & \textbf{26.48} & 61.2 & 3.63 \\
 & 64 & 72.87 & 1.59 & 45.56 & 26.29 & 61.74 & 3.92 \\
 \midrule 
\multirow{4}{*}{WordPiece} & 8 & 98.25 & \textbf{33.33} & 51.84 & 15.71 & 46.47 & 3.79 \\
 & 16 & 99.4 & 22.09 & 59.2 & 13.81 & 59.23 & 2.6 \\
 & 32 & 99.72 & 11.76 & 70.15 & 10.55 & 71.14 & 1.1 \\
 & 64 & \textbf{100} & 4.17 & 84.41 & 5.03 & 81.01 & {\underline{0.44}} \\
 \midrule
\multirow{4}{*}{SP-ULM} & 8 & \textbf{100} & {\underline{0}} & 38.56 & 20.14 & 40.4 & 3.92 \\
 & 16 & \textbf{100} & {\underline{0}} & 42.63 & 23.34 & 53.45 & 3.01 \\
 & 32 & \textbf{100} & {\underline{0}} & 48.05 & 24.38 & 64.91 & 2.83 \\
 & 43 & \textbf{100} & {\underline{0}} & 50.56 & 23.48 & 67.03 & 2.83 \\
 \midrule
\multirow{4}{*}{SP-BPE} & 8 & \textbf{100} & {\underline{0}} & 61.53 & 3.87 & 38.75 & \textbf{7.62} \\
 & 16 & \textbf{100} & {\underline{0}} & 70.01 & 2.66 & 48.77 & 5.6 \\
 & 32 & \textbf{100} & {\underline{0}} & 79.56 & 1.74 & 60.69 & 3.36 \\
 & 64 & \textbf{100} & {\underline{0}} & 89.26 & {\underline{0.79}} & 72.9 & 1.68 \\
 \midrule
\multirow{4}{*}{BPE} & 8 & 79.87 & 26.31 & 66.05 & 4.39 & 40.53 & 6.4 \\
 & 16 & 84.81 & 19.23 & 74.33 & 3.03 & 49.07 & 4.93 \\
 & 32 & 91.76 & 11.68 & 83.23 & 1.82 & 59.87 & 2.97 \\
 & 64 & \textbf{100} & 2.15 & \textbf{91.71} & {\underline{0.87}} & \textbf{82.15} & 0.88 \\
 \midrule
\multirow{4}{*}{Byte-BPE} & 8 & 88.55 & 32.35 & 61.26 & 3.91 & {\underline{38.7}} & 7.51 \\
 & 16 & 92.7 & 24.05 & 69.62 & 2.73 & 48.85 & 5.52 \\
 & 32 & 96.26 & 16.12 & 79.12 & 1.82 & 61.29 & 3.25 \\
 & 64 & 98.01 & 9.77 & 87.67 & 0.95 & 71.75 & 1.91
\end{tabular}%
}
\caption{Absorption Analysis results for the tokenizers evaluated across vocabulary sizes (V) where E, S and P are eclipses, suffixes and prefixes respectively and where Mapping represents the eclipses being merged with their respective subsequent character(s) displayed in Table \ref{table-1-eclipses}.}
\label{table:appendix-absorp}
\end{table}

\subsection{MorphScore Unsuitability}
As mentioned in Section \ref{sec:evaluation}, MorphScore was explored as an evaluation avenue in this research. On examination of the Irish language data included within the framework, we ultimately decided to not implement it. There were three main issues surrounding the data that led to this decision:
\begin{enumerate}
    \item Limited data: 2576 entries in MorphScore for Irish compared to over 35,000 in MoirfEolas
    \item It appears that the morphologically-aligned subwords in MorphScore are based off of a manipulation of the lemma and stem of a word. The stem appears to be a substring common to both the word and the lemma and then the subwords are centered around this stem i.e. `preceding\textunderscore{ }part' is a subword of the characters before the stem and `following\textunderscore{ }part' is a subword of the characters after the stem. This stem creation leads to morphologically-inaccurate subwords in many instances.
    \item Multiple affixes not accounted for i.e. MorphScore only allows for up to three subwords while the Irish morphology explored in this research necessitates up to 4 subwords
\end{enumerate}

Each of the below examples are displayed in Table \ref{table:ms-incorrect}, categorised by the morphological component (eclipses, prefix, suffix) mishandled by the MorphScore scheme. Each of these entries are collected from the Irish portion of MorphScore data, exploring issues 2 and 3 described above. In addition to examining the morphological composition of the words and how the MorphScore subwords do not align, the MoirfEolas subwords are listed. In 5 out of 9 of the below worked examples, even though MoirfEolas contains more subword boundaries than MorphScore, the required number of subwords are equal. As well as this, 3 out of 9 of the below examples need an additional subword and 1 example needs less subwords than the MorphScore entries as per MoirfEolas.

\begin{table}[]
\centering
\resizebox{\columnwidth}{!}{%
\begin{tabular}{ccccccc}
\textbf{Component} & \textbf{Index} & \textbf{Word} & \textbf{Lemma} & \textbf{Stem} & \textbf{preceding\_part} & \textbf{following\_part} \\
\multirow{3}{*}{Eclipses} & 2722 & dtugtar & tabhair & ta & dtug & r \\
 & 11225 & gcinn & ceann & nn & gci & \multicolumn{1}{l}{} \\
 & 30914 & ndeachadar & téigh & h & ndeac & adar \\
\multirow{3}{*}{Prefix} & 2230 & seandaoine & seanduine & seand & \multicolumn{1}{l}{} & aoine \\
 & 8008 & comhionannais & comhionannas & comhionanna & \multicolumn{1}{l}{} & is \\
 & 27027 & mbunscoil & bunscoil & bunscoil & m & \multicolumn{1}{l}{} \\
\multirow{3}{*}{Suffix} & 11331 & osclóimid & oscail & osc & \multicolumn{1}{l}{} & lóimid \\
 & 14158 & t-idirghabhálaí & idirghabhálaí & idirghabhálaí & t- & \multicolumn{1}{l}{} \\
 & 30469 & dtiocfaidh & tar & t & d & iocfaidh
\end{tabular}%
}
\caption{Subset of MorphScore Irish Entries deemed incorrect}
\label{table:ms-incorrect}
\end{table}

\subsubsection{Eclipses}
\textbf{dtugtar}: There is an issue with the lemma -\textgreater{} stem -\textgreater{} subwords calculation causing an incorrect stem of 'ta'. This is a result of being an irregular verb with a lemma quite different to the verb-form used in this case. The incorrect stem is a substring of the verbal suffix 'tar'---the autonomous present tense ending for verbs---added to the word. With the subword boundaries being calculated around the stem, the eclipsis 'd' is absorbed into leading characters of the word and an incorrect `following\textunderscore{ }part' subword created. MorphScore subwords are thus \fcolorbox{black}{cyan}{'dtug'} \fcolorbox{black}{cyan}{'ta'} \fcolorbox{black}{cyan}{'r'} whereas the MoirfEolas subwords are \fcolorbox{black}{lime}{'d'} \fcolorbox{black}{lime}{'tug'} \fcolorbox{black}{lime}{'tar'}.

\textbf{gcinn}: There is an issue with the lemma -\textgreater{} stem -\textgreater{} subword calculation causing an incorrect stem of 'nn' leading to eclipsis absorption into the leading characters of the word `cinn' meaning `choice'. MorphScore subwords are thus
\fcolorbox{black}{cyan}{'gci'} \fcolorbox{black}{cyan}{'nn'} whereas the MoirfEolas subwords are \fcolorbox{black}{lime}{'g'} \fcolorbox{black}{lime}{'cinn'}.

\textbf{ndeachadar}: There is an issue with the lemma -\textgreater{} stem -\textgreater{} subword calculation resulting in a stem of 'h'. This again leads to eclipsis absorption into leading characters and the incorrect `following\textunderscore{ }part' entry of 'adar' as the expected segmentation is surrounding the incorrect stem of 'h'. MorphScore subwords are thus \fcolorbox{black}{cyan}{'ndeac'} \fcolorbox{black}{cyan}{'h'} \fcolorbox{black}{cyan}{'adar'} whereas the MoirfEolas subwords are \fcolorbox{black}{lime}{'n'} \fcolorbox{black}{lime}{'deachadar'}.

\subsubsection{Prefixes}
\textbf{seandaoine}: This compound word contains a prefix 'sean' meaning 'old' and the plural noun 'daoine' meaning 'people' which results in the word 'old people'. Both parts of this word are common in isolation. The lemma -\textgreater{} stem -\textgreater{} subword calculation is incorrect as it doesn't capture this nuance and therefore snowballs into an incorrect entry for 'following\textunderscore{ }part'. MorphScore subwords are thus \fcolorbox{black}{cyan}{'seand'} \fcolorbox{black}{cyan}{'aoine'} whereas the MoirfEolas subwords are \fcolorbox{black}{lime}{'sean'} \fcolorbox{black}{lime}{'daoine'}.

\textbf{comhionannais}: This is a compound word meaning 'equality' with the prefix 'comh' meaning 'joint/mutual' and the adjective 'ionann' meaning 'same/equal'. The lemma -\textgreater{} stem -\textgreater{} subword calculation is incorrect and does not capture this, leading to no entry for 'preceding\textunderscore{ }part'. The addition of the suffix 'ais' to 'ionann' changes the word to the genetive singular case, particularly meaning 'sameness' or 'identity'. This is not captured by MorphScore which has an incorrect entry for 'following\textunderscore{ }part'. MorphScore subwords are thus \fcolorbox{black}{cyan}{'comhionanna'} \fcolorbox{black}{cyan}{'is'} whereas the MoirfEolas subwords are \fcolorbox{black}{lime}{'comh'} \fcolorbox{black}{lime}{'ionann'} \fcolorbox{black}{lime}{'ais'}.

\textbf{mbunscoil}: This compound word means 'primary/elementary school' with the prefix 'bun' meaning 'primary/elementary/basic' and then 'scoil' meaning 'school'. The eclipsis added to the word is segmented correctly however the prefix is not separated. MorphScore subwords are thus \fcolorbox{black}{cyan}{'m'} \fcolorbox{black}{cyan}{'bunscoil'} whereas the MoirfEolas subwords are \fcolorbox{black}{lime}{'m'} \fcolorbox{black}{lime}{'bun'} \fcolorbox{black}{lime}{'scoil'}.

\subsubsection{Suffixes}
\textbf{osclóimid}: The 'l' in this form of the verb 'oscail' meaning 'to open' is still a part of the verb, appearing in almost all conjugations of `to open', and therefore should be included in the stem. The lemma -\textgreater{} stem -\textgreater{} subword calculation does not include it. The verbal suffix 'óimid'---the first person plural future tense ending for verbs that are multi-syllable---should be separated from the root of the verb 'oscl' in this case. Due to the subword boundaries in MorphScore, the `l' is absorbed into the verbal suffix resulting in an incorrect 'following\textunderscore{ }part' entry. MorphScore subwords are thus \fcolorbox{black}{cyan}{'osc'} \fcolorbox{black}{cyan}{'lóimid'} whereas the MoirfEolas subwords are \fcolorbox{black}{lime}{'oscl'} \fcolorbox{black}{lime}{'óimid'}.

\textbf{t-idirghabhálaí}: This is a compound word, containing an eclipsis, prefix, and suffix where 4 subwords are needed to capture the morphology of the word but MorphScore only allows for 3 subwords. The lemma and stem in MorphScore do not adequately capture the compound nature of the word where `idir' is a prefix meaning between and 'gabhál' means `taker' or `catcher'. The word 'gabhál' is also lenited in this case, with an infix of 'h' after the 'g'. Together these words mean 'intercept' or 'intervention'. The addition of the suffix 'aí' personifies the word to mean 'interceptor' or 'intervener'. The eclipses `t-' is added following the determiner `an' meaning `the' and is the only morpheme captured by the MorphScore segmentations. MorphScore subwords are thus \fcolorbox{black}{cyan}{'t-'} \fcolorbox{black}{cyan}{'idirghabhálaí'} whereas the MoirfEolas subwords are \fcolorbox{black}{lime}{'t-'} \fcolorbox{black}{lime}{'idir'} \fcolorbox{black}{lime}{'ghabál'} \fcolorbox{black}{lime}{'aí'}.

\textbf{dtiocfaidh}: This irregular verb meaning `will come' with an eclipses causes issues in the MorphScore segmentations. The lemma -\textgreater{} stem -\textgreater{} subword calculation leads to an incorrect stem of 't' as the lemma is `tar' but the base of the verb in both the conditional and future tense is `tioc', so `t' is the only shared character between both the lemma and stem in MorphScore. While the `preceding\textunderscore{ }part' is correct in this instance, it is only because the segmentation happens around this incorrect stem. Subsequently there is also an incorrect entry for `following\textunderscore{ }part', where the correct entry is the future tense verbal suffix 'faidh'. MorphScore subwords are thus \fcolorbox{black}{cyan}{'d'} \fcolorbox{black}{cyan}{'t'} \fcolorbox{black}{cyan}{'iocfaidh'} whereas the MoirfEolas subwords are \fcolorbox{black}{lime}{'d'} \fcolorbox{black}{lime}{'tioc'} \fcolorbox{black}{lime}{'faidh'}.

\end{document}